\documentclass{article} % For LaTeX2e
\usepackage[preprint]{fastino_paper}
\renewcommand{\headrulewidth}{0pt}

\usepackage{lmodern}
\usepackage{microtype}
\usepackage{hyperref}
\usepackage{url}
\usepackage[T1]{fontenc}
\usepackage{booktabs}
\usepackage{graphicx}
\usepackage{multirow}
\usepackage[table]{xcolor}
\usepackage{siunitx}
\usepackage[most]{tcolorbox}
\usepackage{setspace}
\usepackage{hyphenat}
\usepackage{placeins}
\usepackage{siunitx}
\usepackage{amsmath}
\usepackage{amssymb}
\usepackage{enumitem}
\usepackage{array}
\usepackage{pifont}
\usepackage{textcomp}
\usepackage{lineno}
\usepackage{listings}

\definecolor{fastinoteal}{HTML}{000000}
\definecolor{fastinodark}{HTML}{000000}
\definecolor{fastinolight}{HTML}{F5F7FA}
\definecolor{fastinogray}{HTML}{4B5563}
\definecolor{darkblue}{rgb}{0, 0, 0.5}
\definecolor{piiHeader}{HTML}{FF7345}
\definecolor{piiGroup}{HTML}{0C0608}
\definecolor{piiRowA}{HTML}{FDF2EC}
\definecolor{piiRowB}{HTML}{F2ECE9}
\definecolor{piiFrame}{HTML}{F8D5A5}

\definecolor{fastinoOrange}{HTML}{FF7345}
\definecolor{fastinoCream}{HTML}{F2ECE9}
\definecolor{fastinoInk}{HTML}{0C0608}

\hypersetup{
  colorlinks=true,
  citecolor=darkblue,
  linkcolor=darkblue,
  urlcolor=darkblue
}

\definecolor{pycharmBg}{HTML}{FFFFFF}
\definecolor{pycharmFg}{HTML}{000000}
\definecolor{pycharmKeyword}{HTML}{0033B3}
\definecolor{pycharmString}{HTML}{067D17}
\definecolor{pycharmComment}{HTML}{8C8C8C}
\definecolor{pycharmNumber}{HTML}{1750EB}
\definecolor{pycharmBuiltin}{HTML}{8000FF}
\definecolor{pycharmFrame}{HTML}{E0E0E0}
\definecolor{pycharmLineNo}{HTML}{999999}

\lstdefinestyle{pythoncompact}{
  language=Python,
  basicstyle=\ttfamily\footnotesize\color{pycharmFg},
  keywordstyle=\bfseries\color{pycharmKeyword},
  commentstyle=\itshape\color{pycharmComment},
  stringstyle=\color{pycharmString},
  identifierstyle=\color{pycharmFg},
  emph={self,True,False,None},
  emphstyle=\bfseries\color{pycharmKeyword},
  emph={[2]print,len,range,open,list,dict,tuple,set,int,str,float,bool,type,isinstance},
  emphstyle={[2]\color{pycharmBuiltin}},
  captionpos=b,
  backgroundcolor=\color{pycharmBg},
  showstringspaces=false,
  frame=single,
  framerule=0.4pt,
  rulecolor=\color{pycharmFrame},
  breaklines=true,
  columns=fullflexible,
  keepspaces=true,
  framesep=8pt,
  numbers=left,
  numberstyle=\sffamily\tiny\color{pycharmLineNo},
  numbersep=8pt,
  xleftmargin=10pt,
  xrightmargin=4pt,
  aboveskip=6pt,
  belowskip=2pt
}

\begin{document}

% --- Fastino-style title box ---
\tcbset{enhanced,frame hidden}
\tcbset{left=0.6cm, right=0.6cm, top=0.2cm, bottom=0.2cm}
\tcbset{arc=6pt}
\tcbset{colback=fastinolight}
\tcbset{before skip=0pt}
\tcbset{grow to left by=1.5pt, grow to right by=1.5pt}
\tcbset{borderline west={3pt}{0pt}{fastinoteal}}
\tcbset{overlay={\node[
    anchor=south east,
    at=(frame.south east),
    xshift=-0.5cm,
    yshift=0.45cm] {\includegraphics[width=3cm]{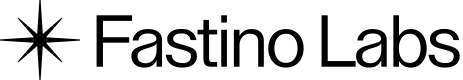}};}}
\begin{tcolorbox}
  \setlength{\parindent}{0cm}
  \setlength{\parskip}{0.15cm}
  {
    \setlength{\parskip}{0cm}
    \raggedright
    \nohyphens
    {
      \setstretch{1.618}
            {\LARGE\sffamily\bfseries\color{fastinodark} GLiNER2-PII: A Multilingual Model for Personally Identifiable Information Extraction}\par\par
    }
    \vskip 0.35cm
    {\normalsize\sffamily\bfseries Urchade Zaratiana}\hspace{0.8em}%
    {\normalsize\sffamily\bfseries Ash Lewis}\hspace{0.8em}%
    {\normalsize\sffamily\bfseries George Hurn-Maloney}\footnotemark\par
    \vskip 0.15cm

    {\normalsize\sffamily\bfseries\color{fastinogray} Fastino Labs}\par
    \vskip 0.08cm
  }
  {\color{fastinodark}
  \textbf{Abstract.}\quad Reliable detection of personally identifiable information (PII) is increasingly important across modern data-processing systems, yet the task remains difficult: PII spans are heterogeneous, locale-dependent, context-sensitive, and often embedded in noisy or semi-structured documents. We present GLiNER2-PII, a small 0.3B-parameter model adapted from GLiNER2 and designed to recognize a broad taxonomy of 42 PII entity types at character-span resolution. Training such systems, however, is constrained by the scarcity of shareable annotated data and the privacy risks associated with collecting real PII at scale. To address this challenge, we construct a multilingual synthetic corpus of 4,910 annotated texts using a constraint-driven generation pipeline that produces diverse, realistic examples across languages, domains, formats, and entity distributions. On the challenging SPY benchmark, GLiNER2-PII achieves the highest span-level F1 among five compared systems, including OpenAI Privacy Filter and three GLiNER-based detectors. We publicly release the model on Hugging Face to support further research and practical deployment of open PII detection systems.
  }\par
  \vskip 0.15cm
  {
    \setlength{\parskip}{0cm}
    {\small {\sffamily \bfseries GitHub:} \href{https://github.com/fastino-ai/GLiNER2}{\texttt{github.com/fastino-ai/GLiNER2}}}\par
    {\small {\sffamily \bfseries Model:} \href{https://huggingface.co/fastino/gliner2-privacy-filter-PII-multi}{\texttt{hf.co/fastino/gliner2-privacy-filter-PII-multi}}}\par
  }
\end{tcolorbox}
\footnotetext{Correspondence to George Hurn-Maloney. Email: \href{mailto:g@fastino.ai}{\texttt{g@fastino.ai}}.}
\tcbset{reset}
\FloatBarrier

\section{Introduction}

Personally identifiable information (PII) is pervasive in modern digital systems, appearing in customer communications, support tickets, CRM records, financial and healthcare documents, system logs, and authentication data. As organisations increasingly route this text through automated pipelines for analytics, search, model development, and operational tooling, regulatory frameworks such as GDPR and CCPA make the reliable detection and removal of sensitive information a prerequisite rather than an optional safeguard. Deployed systems must therefore operate directly on unstructured, often noisy text and return precise character-level spans that downstream components can mask, audit, or route accordingly.

Reliable PII detection is difficult for several reasons. Entity formats vary by locale (phone numbers, tax IDs, addresses, IBANs, passports), many values are ambiguous without surrounding context, and real documents often contain signatures, quoted replies, forms, logs, and multilingual fragments where PII is nested or interleaved with non-sensitive text. At the same time, the detection system must balance precision and recall carefully: missed spans create privacy and compliance risk, while overly aggressive masking degrades data utility for analytics, support automation, search, and model training.

Existing approaches generally fall into two broad families. Token-classification models with constrained decoding \citep{openai_privacy_filter_2026} are efficient, but are typically designed around predefined label schemas that can be difficult to adapt or extend. In contrast, label-conditioned span extractors such as the GLiNER family \citep{zaratiana-etal-2024-gliner} treat target labels as inputs, allowing the same architecture to support a wide range of extraction tasks. This flexibility is particularly important for PII detection, where practical deployments often require distinctions between closely related entity types and support for organization- or jurisdiction-specific policies. For example, different categories of dates or identifiers may need to be retained, masked, or audited differently depending on downstream requirements.

This report presents GLiNER2-PII, a PII detection and masking model fine-tuned from GLiNER2~\citep{zaratiana-etal-2025-gliner2, zaratiana2026gliguardschemaconditionedclassificationllm}. Our contributions are: (i)~a detector covering a fine-grained inventory of 42~entity types across seven categories: personal, contact, governmental, financial, digital identity, credential, and calendar date; (ii)~a constraint-driven LLM synthesis procedure, built on the data-generation framework of Pioneer Agent~\citep{atreja2026pioneeragentcontinualimprovement}, that produces a multilingual corpus of 4{,}910~annotated PII examples; and (iii)~an evaluation on the SPY benchmark~\citep{savkin-etal-2025-spy} on which GLiNER2-PII achieves the highest span-level F1 among five compared systems.

\begin{figure}
    \centering
    \includegraphics[width=0.98\linewidth]{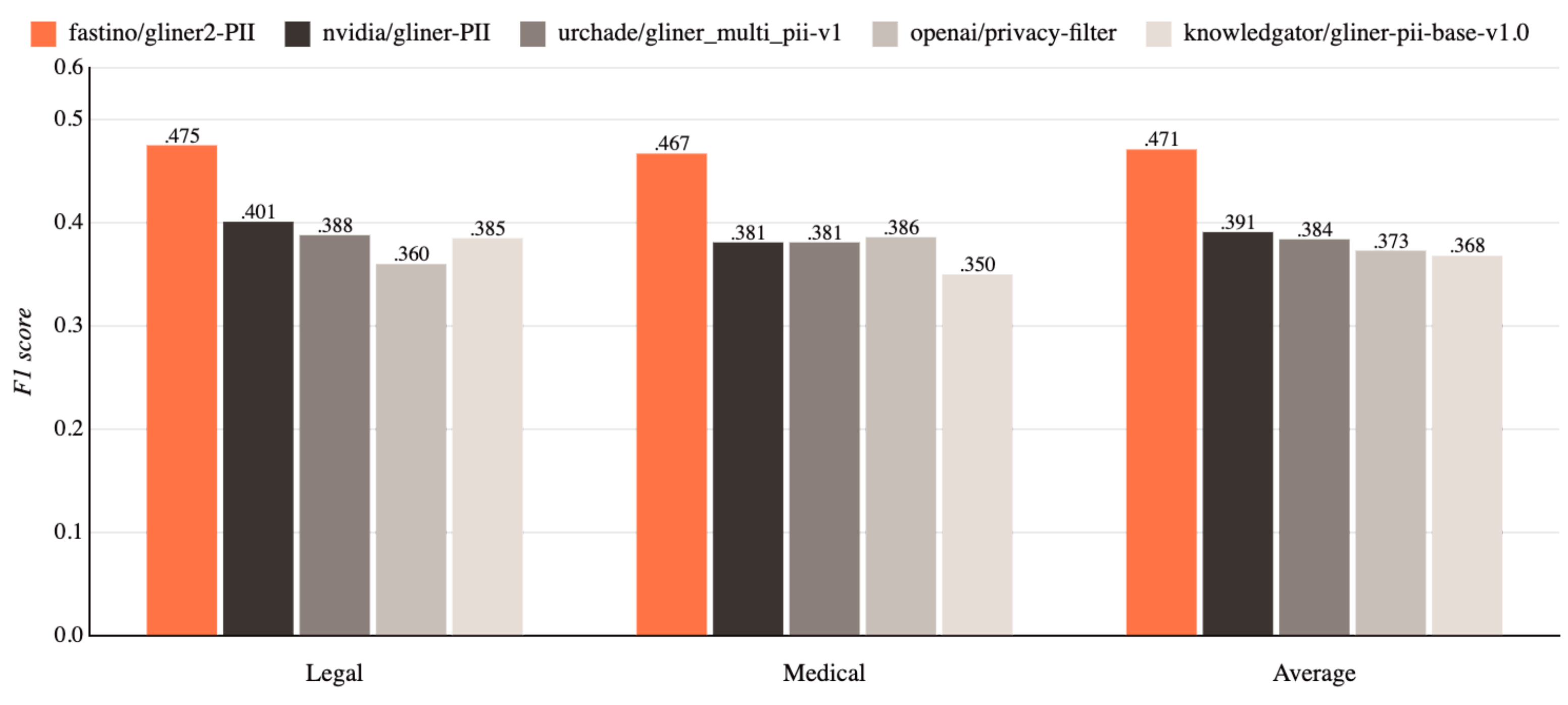}
    \caption{\textbf{Span-level F1 on the SPY benchmark} \citep{savkin-etal-2025-spy}. GLiNER2-PII outperforms the three baselines across both domains.}
    \label{fig:placeholder}
\end{figure}

\section{Method}

We cast PII detection as schema-based entity extraction. Given an input text $x$ and a schema $\mathcal{Y} = \{(y_i, d_i)\}_{i=1}^{M}$ of target entity types with optional natural-language descriptions, the model returns extracted spans and their types. A downstream redaction system then masks the matched substrings.

We build on GLiNER2~\citep{zaratiana-etal-2025-gliner2}, a compact 0.3B-parameter unified information extraction model that supports entity extraction, classification, structured extraction, and relation extraction within a common schema interface. Because the model conditions on a set of target labels at inference time, the same architecture can serve different PII schemas without modification. In this report, we fine-tune the model for entity extraction over a 42-label PII schema, producing exact character spans suited to downstream masking.

\begin{lstlisting}[style=pythoncompact,caption={Minimal GLiNER2-PII inference example.},label={lst:inference}]
from gliner2 import GLiNER2

model = GLiNER2.from_pretrained("fastino/gliner2-privacy-filter-PII-multi")

text = "Email john.smith@acme.com or call +1 415 555 0199."
labels = ["email", "phone_number", "person"]

result = model.extract_entities(
    text,
    labels,
    threshold=0.5,
    include_confidence=True,
    include_spans=True,
)
\end{lstlisting}

\paragraph{Inference example.} Listing~\ref{lst:inference} shows a minimal inference workflow. The API loads the fine-tuned checkpoint, defines a target label set, and returns extracted entities with confidence scores and character spans.

\section{Data Generation}
\label{sec:data_generation}

Collecting naturally occurring PII at scale is difficult, since the most realistic examples are also the most sensitive and the hardest to share. We therefore construct a synthetic corpus using the constraint-driven data-generation framework introduced for Pioneer Agent~\citep{atreja2026pioneeragentcontinualimprovement}. Given a natural-language description of the extraction task, the framework automatically derives a set of sampling constraints covering label composition, document format, and language, and uses them to condition large frontier decoder to produce diverse, schema-compliant annotated examples.

\paragraph{Label inventory.} The schema covers 42~entity types organised into seven groups (Table~\ref{tab:pii_labels}). Coarse labels such as \texttt{person} or \texttt{payment\_card} support broad masking policies, while fine-grained subtypes such as \texttt{first\_name}, \texttt{card\_number}, and \texttt{card\_cvv} enable precise redaction and policy-specific routing. Annotations cover the exact substring of each PII value and exclude surrounding words, punctuation, and field labels, except where these are themselves part of the value. Crucially, the inventory permits nested entities: a single text region may carry multiple overlapping labels at different levels of granularity. For example, a \texttt{full\_name} span may contain inner \texttt{first\_name} and \texttt{last\_name} spans, and a URL annotated as a sensitive endpoint may contain an inner \texttt{access\_token} or \texttt{api\_key}. This design lets downstream redaction policies operate at whichever level of granularity they require, either by masking the entire outer span when conservative behaviour is needed or by surgically redacting only the inner credential when the surrounding context must be preserved.

\begin{table*}[t]
\centering
\small
\renewcommand{\arraystretch}{1.2}
\setlength{\tabcolsep}{6pt}

\begin{tabular}{
@{}
>{\bfseries\raggedright\arraybackslash}p{0.23\linewidth}
>{\raggedright\arraybackslash}p{0.72\linewidth}
@{}
}
\toprule
\textbf{Semantic group} & \textbf{PII label types} \\
\midrule
\rowcolor{gray!6}
Person / identity
&
\texttt{person},
\texttt{full\_name},
\texttt{first\_name},
\texttt{middle\_name},
\texttt{last\_name},
\texttt{date\_of\_birth}
\\
Contact / location
&
\texttt{email},
\texttt{phone\_number},
\texttt{address},
\texttt{street\_address},
\texttt{city},
\texttt{state\_or\_region},
\texttt{postal\_code},
\texttt{country}
\\
\rowcolor{gray!6}
Government / tax identifiers
&
\texttt{government\_id},
\texttt{national\_id\_number},
\texttt{passport\_number},
\texttt{drivers\_license\_number},
\texttt{license\_number},
\texttt{tax\_id},
\texttt{tax\_number}
\\
Banking / payment
&
\texttt{bank\_account},
\texttt{account\_number},
\texttt{routing\_number},
\texttt{iban},
\texttt{payment\_card},
\texttt{card\_number},
\texttt{card\_expiry},
\texttt{card\_cvv}
\\
\rowcolor{gray!6}
Digital identity
&
\texttt{username},
\texttt{ip\_address},
\texttt{account\_id},
\texttt{sensitive\_account\_id}
\\
Secrets / credentials
&
\texttt{password},
\texttt{secret},
\texttt{api\_key},
\texttt{access\_token},
\texttt{recovery\_code}
\\
\rowcolor{gray!6}
Sensitive dates
&
\texttt{sensitive\_date},
\texttt{document\_date},
\texttt{expiration\_date},
\texttt{transaction\_date}
\\
\bottomrule
\end{tabular}

\caption{
\textbf{PII label inventory.}
The model covers 42 fine-grained PII entity types grouped into seven semantic categories spanning identity, contact information, financial data, credentials, and sensitive temporal attributes.
}
\label{tab:pii_labels}

\end{table*}

\paragraph{Synthetic data pipeline.} The framework takes two inputs: the PII label inventory and a natural-language description of the extraction task. It uses these to build two complementary sets of constraints. From the label inventory, it derives \emph{programmatic constraints} that control which labels appear in each example, including entity-type counts, label-exclusion subsets, and at-least-one requirements. From the task description, it infers \emph{diversity constraints} that control the surface form of each example, including document type, locale, register, and tone. To generate a single example, the framework samples a subset of these constraints, formats them into a prompt, and queries large frontier decoder (temperature~0.01) to produce the text together with its span-level annotations. Running this loop yields a corpus spanning chat logs, support tickets, CRM notes, KYC forms, invoices, medical records, and credential files, in English, French, Spanish, German, Italian, Portuguese, and Dutch, with occasional mixed-language passages.

\section{Evaluation Setup}

\paragraph{Benchmark.} We evaluate on SPY (Synthetic PII Yesterday)~\cite{savkin-etal-2025-spy}, which contains two domain-specific subsets, \emph{Legal Questions} (100 documents from legal Q\&A forums) and \emph{Medical Consultations} (100 documents from medical transcripts), annotated with seven PII types: \textsc{name}, \textsc{address}, \textsc{email}, \textsc{phone\_num}, \textsc{id\_num}, \textsc{url}, \textsc{username}. We chose SPY because it provides recent, naturally formatted text that is better suited for measuring out-of-distribution generalization. We deliberately exclude datasets such as \texttt{ai4privacy/pii-masking-300k}, since several publicly available PII models were trained on its training split, making it difficult to isolate true OOD performance.

\paragraph{Baselines.} We compare against four publicly available PII detectors representing both token-classification and label-conditioned extraction approaches. \textit{OpenAI Privacy Filter}~\citep{openai_privacy_filter_2026} is a bidirectional token classifier with BIOES decoding over 8 coarse-grained categories. \textit{NVIDIA GLiNER PII}~\citep{nvidia_gliner_pii_2026} is a GLiNER-based model optimized for practical PII extraction with a relatively narrow label schema. \texttt{urchade/gliner\_multi\_pii-v1}~\citep{zaratiana_gliner_multi_pii_v1_2024} is a multilingual GLiNER fine-tune supporting a compact set of entity types across languages. Finally, \texttt{knowledgator/gliner-pii-base-v1.0}~\citep{knowledgator_gliner_pii_collection} is another GLiNER-based PII detector trained on a restricted inventory of PII categories. 

\paragraph{Label mapping.} Because each model uses a different internal label set, we apply a deterministic label mapping for every model so that its predictions align with SPY's seven evaluation categories. The mapping procedure is identical across all systems to ensure a fair comparison.

\paragraph{Metrics.} We report span-level precision, recall, and F1 under exact-match evaluation: a prediction is counted as correct only when both its label type and character boundaries match a gold span. We place particular emphasis on recall, since in redaction settings false negatives leave sensitive information unmasked.

\begin{table}[t]
\centering
\small
\setlength{\tabcolsep}{5pt}
\renewcommand{\arraystretch}{1.15}

\begin{tabular}{
l
S[table-format=1.3]
S[table-format=1.3]
S[table-format=1.3]
S[table-format=1.3]
S[table-format=1.3]
S[table-format=1.3]
S[table-format=1.3]
}
\toprule
& \multicolumn{3}{c}{\textbf{Legal}}
& \multicolumn{3}{c}{\textbf{Medical}}
& \multicolumn{1}{c}{\multirow{2}{*}{\textbf{Avg.\ F1}}} \\
\cmidrule(lr){2-4}
\cmidrule(lr){5-7}
\textbf{Model}
& {\textbf{P}}
& {\textbf{R}}
& {\textbf{F1}}
& {\textbf{P}}
& {\textbf{R}}
& {\textbf{F1}}
& \\
\midrule
nvidia/gliner-PII
& {\cellcolor{yellow!15}0.374}
& 0.431
& {\cellcolor{yellow!15}0.401}
& 0.341
& 0.431
& {\cellcolor{yellow!15}0.381}
& {\cellcolor{yellow!15}0.391} \\
urchade/gliner\_multi\_pii-v1
& {\cellcolor{green!20}\bfseries 0.522}
& 0.308
& 0.388
& {\cellcolor{green!20}\bfseries 0.483}
& 0.314
& {\cellcolor{yellow!15}0.381}
& 0.384 \\
openai/privacy-filter
& 0.250
& {\cellcolor{yellow!15}0.640}
& 0.360
& 0.271
& {\cellcolor{yellow!15}0.671}
& {\cellcolor{yellow!15}0.386}
& 0.373 \\
knowledgator/gliner-pii-base-v1.0
& {\cellcolor{yellow!15}0.398}
& {\cellcolor{yellow!15}0.372}
& 0.385
& {\cellcolor{yellow!15}0.389}
& {\cellcolor{yellow!15}0.319}
& 0.350
& 0.368 \\
\midrule
\textbf{fastino/gliner2-PII}
& 0.354
& {\cellcolor{green!20}\bfseries 0.722}
& {\cellcolor{green!20}\bfseries 0.475}
& 0.355
& {\cellcolor{green!20}\bfseries 0.681}
& {\cellcolor{green!20}\bfseries 0.467}
& {\cellcolor{green!20}\bfseries 0.471} \\
\bottomrule
\end{tabular}

\caption{
\textbf{Span-level PII detection performance on SPY}~\cite{savkin-etal-2025-spy}.
Reported metrics are exact-match precision (P), recall (R), and F1.
Best results in each column are shown in \textbf{bold}.
}
\label{tab:main_results}

\end{table}

\section{Results}
\label{sec:results}

Table~\ref{tab:main_results} summarizes the main results. GLiNER2-PII achieves the highest exact-match F1 on both the legal and medical subsets, as well as the best overall average.

\paragraph{Recall.} GLiNER2-PII obtains recall scores of 0.722 on the legal subset and 0.681 on the medical subset, substantially outperforming the other GLiNER variants (0.308--0.431). OpenAI Privacy Filter reaches similarly high recall (0.640--0.671), but with much lower precision (0.250--0.271), leading to a larger number of false positives per document. For redaction settings, where missing a PII span is often more costly than over-redaction, the higher recall of GLiNER2-PII is particularly important.

\paragraph{Precision and recall trade-off.} \texttt{urchade/gliner\_multi\_pii-v1} achieves the highest precision (0.483--0.522), but does so by predicting more conservatively, resulting in the lowest recall (0.308--0.314). NVIDIA GLiNER-PII and knowledgator/gliner-pii-base-v1.0 show a more balanced trade-off between precision and recall. GLiNER2-PII favors recall while maintaining competitive precision, making it better suited for practical redaction workflows. Performance is generally consistent across the legal and medical domains for all evaluated systems.

\section{Discussion and Limitations}
\label{sec:discussion}

Despite being trained entirely on large frontier decoder-generated synthetic data, GLiNER2-PII achieves the highest F1 on both evaluation sets, which are drawn from naturally occurring text. This suggests that controlled synthetic generation can provide enough diversity in formatting, writing style, and entity composition to support transfer across domains. We also observe that using a fine-grained inventory of 42 labels may help the model learn stronger representations of broader entity categories. In particular, the largest gains over the other GLiNER variants appear on \textsc{name} entities.

Precision remains the main area for improvement. Error analysis shows that the model tends to over-predict \textsc{name} entities, sometimes confusing personal names with common nouns, organization names, or product names. Techniques such as label-specific thresholds, lightweight filtering, or calibration on a small validation set could likely improve precision without substantially reducing recall.

Some limitations should also be noted. The evaluation covers only legal and medical documents, and the training data are entirely synthetic and have not been validated by human annotators. At the same time, the strong performance on naturally occurring text suggests that the Pioneer-based synthetic generation pipeline \citep{atreja2026pioneeragentcontinualimprovement} is capable of producing training data that transfer effectively to real-world PII detection tasks.

Overall, the results indicate that combining diverse synthetic training data with a broad label inventory can produce effective PII detectors, even for naturally occurring text from domains not seen during training. Future work includes human-validated fine-tuning, extending coverage to additional locales and languages, broader multilingual evaluation, and end-to-end benchmarking of redaction systems with both accuracy and efficiency metrics.

\bibliographystyle{fastino_paper}
\bibliography{fastino_references}

\end{document}